\documentclass[10pt,twocolumn,letterpaper]{article}

\usepackage{iccv}
\usepackage{times}
\usepackage{epsfig}
\usepackage{graphicx}
\usepackage{amsmath}
\usepackage{amssymb}
\usepackage{multirow}
\usepackage{float}
\usepackage{caption}
\usepackage{authblk}
\usepackage[breaklinks=true,bookmarks=false]{hyperref}

\iccvfinalcopy 

\def\iccvPaperID{4} 

\ificcvfinal\fi

\begin{document}

\title{Unknown Identity Rejection Loss: Utilizing Unlabeled Data for Face Recognition}

\author[1]{Haiming Yu}
\author[1]{Yin Fan}
\author[2]{Keyu Chen}
\author[1]{He Yan}
\author[1]{Xiangju Lu}
\author[1]{Junhui Liu}
\author[1]{Danming Xie}

\affil[1]{iQIYI Inc, China \authorcr
  \{\tt yuhaiming, fanyin, yanhe, luxiangju, liujunhui, xiedanming\}@qiyi.com}
\affil[2]{Beijing Institute of Technology, China \authorcr
  \{\tt ckybit\}@gmail.com}

\maketitle
\ificcvfinal\thispagestyle{empty}\fi

\begin{abstract}
Face recognition has advanced considerably with the availability of large-scale labeled datasets. However, how to further improve the performance with the easily accessible unlabeled dataset remains a challenge. In this paper, we propose the novel Unknown Identity Rejection (UIR) loss to utilize the unlabeled data. We categorize identities in unconstrained environment into the known set and the unknown set. The former corresponds to the identities that appear in the labeled training dataset while the latter is its complementary set. Besides training the model to accurately classify the known identities, we also force the model to reject unknown identities provided by the unlabeled dataset via our proposed UIR loss. In order to `reject' faces of unknown identities, centers of the known identities are forced to keep enough margin from centers of unknown identities which are assumed to be approximated by the features of their samples. By this means, the discriminativeness of the face representations can be enhanced. Experimental results demonstrate that our approach can provide obvious performance improvement by utilizing the unlabeled data.
\end{abstract}


\section{Introduction}

In recent years, face recognition has been extensively studied and applied both in academic and industrial communities. This is largely imputed to the high capacity of convolutional neural networks (CNNs) for learning discriminative representations in image classification. Current state-of-the-art approaches for face identification, verification and clustering mainly take the advantage of the learned face representation, which is obtained by mapping the face image into a lower dimensional embedding through a deep convolutional network. Generally speaking, there exist three factors that affect the discriminativeness of the learned face representation. The first one is the network structure. Well-designed backbone networks (e.g. \cite{resnet} \cite{inceptionv4}) provide high capacity for learning the features. Secondly, the loss function (e.g. \cite{centerloss} \cite{arcface} \cite{additivemargin}) for training also plays an important role, determining how images are mapped into the feature space. Last but not least is the dataset deployed for training the network. Empirically a large-scale dataset of good quality (e.g. \cite{msceleb1m} \cite{vggface2} \cite{iqiyivid}) can sometimes compensate the withdraws of the backbone network and the loss function. Big data has been considered one of the most significant factors in boosting the performance of deep CNN models.

In recent years, several large-scale face recognition datasets have been made available to the public (e.g. \cite{vggface2} \cite{vggface} \cite{msceleb1m} \cite{casiawebface} \cite{megaface}). Among them,  MS-Celeb-1M \cite{msceleb1m} is currently the largest static image dataset, which contains more than 10K identities and 10m of images. The recently released iQIYI-VID \cite{iqiyivid} contains 5K identities and 600K video clips. However, the process of annotating these large-scale datasets is complicated. On the other hand, obtaining unlabeled face images is much easier. For example, a web crawler equipped with a face detector can produce large amounts of in-the-wild face images. Meanwhile, face recognition is not a closed-set classification problem. The number of identities in an unconstrained environment can be unlimited and hundreds of thousands of identities within an annotated dataset can not represent the vast variety of identities in the world. Supplementing the number of identities with unlabeled data for learning discriminative representation is thus worth considering. Therefore, how to leverage these unlabeled data has become a challenging issue in face recognition.

\begin{figure}[h]
\includegraphics[width=8.4cm, height=4cm]{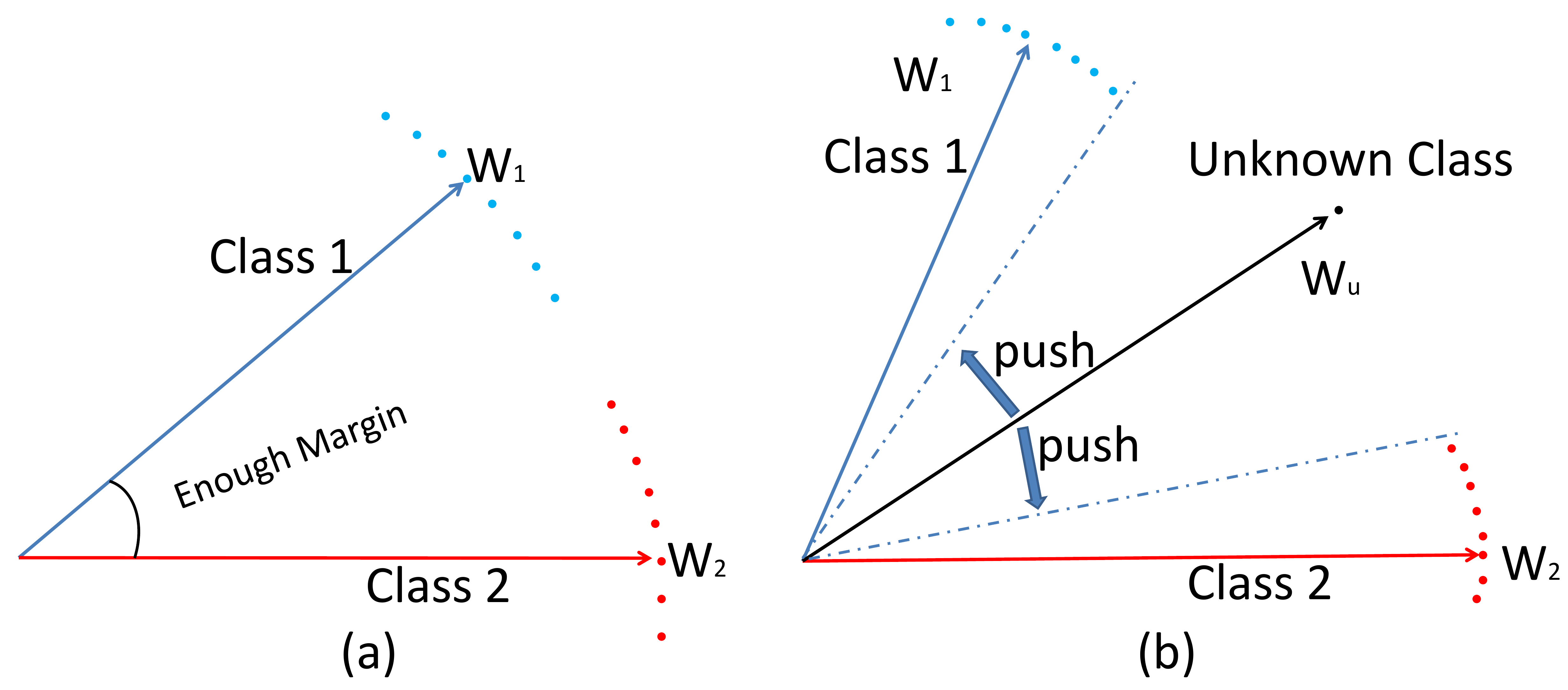}
\centering
\caption{\textmd{Geometrical interpretation of the normalized features. (a): Features and their centers ($W_1$ and $W_2$) of two identities that have enough margin between them. (b): When a sample with an unknown identity ($W_u$) is encountered, the decision margins become insufficient. Therefore extra space must be allocated to the sample with unknown identity, rendering the centers distributed as sparsely as possible. Best viewed in color.}}
\label{fig:figure1}
\end{figure}

Although this problem seems reminiscent to traditional semi-supervised learning (SSL), utilizing unlabeled data for learning discriminative face representation differs from traditional SSL approaches due to the following two challenges. Firstly, web-crawled face images can suffer severely from long-tailed distribution. Given an identity, there may exist only one corresponding image. Therefore, assigning pseudo-labels by clustering the unlabeled images may not provide accurate results. Meanwhile, the overlap between the identities of labeled and unlabeled dataset is never guaranteed. Identities of both the labeled and unlabeled datasets usually have very small overlap. Hence, label propagation \cite{labelpropagation} can not be applied here.

To overcome the aforementioned challenges, we probe into approaches for leveraging unlabeled data. Currently, a discriminative feature extractor is usually trained in a classification manner. Given face images distributed in many different identities, a CNN model is trained to classify an image into its corresponding identity. However, face recognition is an open-set problem \cite{opensetdn} \cite{opensetfr}. Under unconstrained environment, there exist many identities that the classification model did not encounter when it was trained (\textit{i.e.} identities that are not included in the training dataset, whose centers therefore are not represented by the row vectors of the last fully-connected layer). Ideally at query time, when such a sample appears, the probabilities produced by the classifier (usually a fully-connected layer followed by the softmax function) should be as small as possible so that this face can be rejected by any of the categories by a predefined threshold. 

Based on this principle, we designed a loss function, called unknown identity rejection (UIR) loss function to utilize the unlabeled data. We categorize the identities in an open environment into two sets. The first set is the identities that appear in the labeled dataset and represented in the weights of the fully-connected layer, denoted as $\mathcal{S}$. We call them the known identities. Another set is the identities that do not exist in the labeled dataset, denoted as $\mathcal{U}$. We call them unknown identities. Note that $\mathcal{S}\cap \mathcal{U} = \emptyset$. Given the fact that the unlabeled dataset has negligible identity overlap with the labeled dataset, we approximate the set $\mathcal{U}$ with our unlabeled dataset. We feed these data into the classification model and train the model to `reject' them. In this manner, the unlabeled data can be utilized. From the perspective of geometrical interpretation, rejecting unknown identities can also enlarge the inter-class distance. The feature of an `unknown' sample approximates the center of its own identity, therefore extra space must be allocated to its corresponding identity, rendering the distribution of the centers sparse (see Fig. \ref{fig:figure1}).

In general, the main contribution of this paper is the novel Unknown Identity Rejection loss function and its advantages can be summarized as follows: 

a). It can utilize the unlabeled dataset to implicitly increase the number of classes in the classification task hence approximating the real-world distribution of identities in the open environment. This not only enhances the discrininativeness of the learned representation but also benefits the performance of open-set recognition.

b). It is robust to unbalanced distribution over identities (e.g. long-tailed distribution \cite{rangeloss}). Even many of the identities have only one or two images in the unlabeled dataset, our proposed UIR loss is still able to utilize them without attenuating the performance. 

c). UIR loss is easy to implement and can be applied to various tasks without any modification. 

\section{Related Work}

\subsection{Deep Face Recognition}

The advancement in face recogintion in recent years has been focusing on designing various loss functions to optimize the distance metrics among the extraced face features. These works can be classified into two categories. One category is loss functions that optimize the Euclidean distance. Examples include contrastive loss \cite{contrastiveloss}, range loss \cite{rangeloss}, triplet loss \cite{tripletloss}, center loss \cite{centerloss} and marginal loss \cite{marginalloss}. They cluster samples from the same identity by minimizing the intra-class Euclidean distances. The other category is cosine-similarity-based loss functions, e.g. L-Softmax \cite{lsoftmax}, SphereFace \cite{sphereface}, additive cosine margin \cite{additivemargin}, ArcFace \cite{arcface}. These works studied different approaches to compress the angular margin of samples from the same identity. 

However, all these proposed loss functions rely heavily on annotations. Labels in the training dataset serve as supervisory signals to distinguish between intra-class and inter-class samples. Our proposed UIR loss function, on the other hand, is constructed without the label information, while at the same time is able to compress the decision boundary of diverse identities. 

\subsection{Semi-supervised Deep Learning}

To our knowledge, we classify semi-supervised deep learning approaches into two categories. The first category is consistency-based approaches and the second category relies heavily on clustering. 

\textbf{Consistency-based approaches}: These algorithms are based on the fact that, given an object, one should draw the same conclusion of what it is when viewing it from different perspectives. Examples of this approaches include $\Pi$-model and Temporal Ensembling \cite{temporalensembling}, Mean Teacher \cite{meanteacher}, Deep Co-Training \cite{deepcotraining}, etc. They construct different ensemble models (either explicit or implicit) to provide different `views' for the unlabeled data and optimize various versions of consistency loss functions to utilize the unlabeled data. The drawback of these approaches is that the implementation can be very complicated when handling members of the overall ensemble model, especially when the ensemble model is constructed implicitly.

\textbf{Clustering-based approaches}: Clustering is a very commonly used unsupervised learning approach. When handling unlabeled data, one can first cluster these data, assigning fake labels to these data and then use these pseudo-labels to train a classification model. Works like \cite{consensus} \cite{transductive} have applied clustering to deep face recognition. However, as mentioned above, web-crawled unlabeled data face the problem of diverse identity distribution. When there exists only one image within a class, the clustering result can be highly unreliable. In the aforementioned two works, the experiments were done by separating the off-the-shelf annotated dataset into two parts and treat one of them as unlabeled. This practice has actually avoided facing the problem of diverse identity distribution. Moreover, when the amount of unlabeled data is huge, clustering so many images can be extremely computationally expensive. Based on these, we do not take it for granted that clustering is a good solution to semi-supervised face recognition. 

\begin{figure*}
\includegraphics[width=18cm, height=5.85cm]{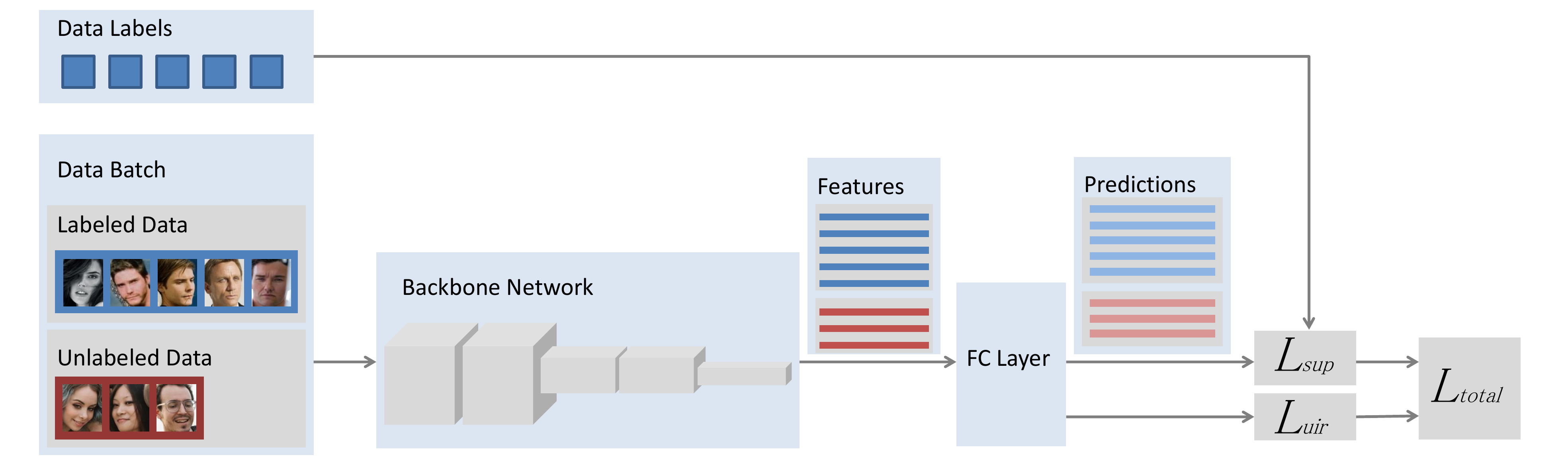}
\caption{\textmd{Overview of the semi-supervised training strategy. A data batch contains both labeled and unlabeled samples. Their features are extracted via a backbone network. Then the predictions of labeled and unlabeled data are used to separately calculate the fully-supervised loss and the UIR loss. The total loss is a weighted sum of the two losses. Best viewed in color.}}
\label{fig:figure2}
\end{figure*}

\section{Approach}

In this section, we first illustrate the formulation of our proposed loss function, and then we discuss some details on the implementation. 

\subsection{Unknown Identity Rejection Loss Function}

Given a CNN classification model, we denote the output (softmax-normalized) of the last fully-connected layer as $p_1, p_2, ..., p_n$, where $n$ is the number of classes. Generally the classification result is $i$ such that $i=\underset{i}{\mathrm{argmax}}\ p_i$. $p_1, p_2, ..., p_n$ indicate how close a sample is to the centers of different identities. However, given a sample $x$ with an unknown identity, since this identity has never appeared in the closed-set classification and belongs to none of the classes, ideally all $p_1, p_2, ..., p_n$ must be as small as possible so that it can be rejected by a predefined threshold. Based on the fact that $p_1, p_2, ..., p_n$ satisfy $p_1+p_2+...+p_n=1$, we formulate the problem as:
\begin{align*}
minimize\ & p_1, p_2, ..., p_n
\\
s.t.\ & \sum_{i=1}^{n} p_i = 1
\end{align*}

This is a multi-objective constrained minimization problem, which does not have one single solution. However, based on the fact that this identity does not belong to any of the known classes, we hope that the probability distribution ($p_1, p_2, ..., p_n$) does not bias to any of the classes. Hence, a plausible solution to this problem can be formulated as
 
\begin{equation*}
p_1=p_2=...=p_n=\frac{1}{n}
\end{equation*}

We notice that this solution is also the optima for the problem
\begin{align*}
maximize\ & p_1 \cdot p_2 \cdot ... \cdot p_n
\\
s.t. & \sum_{i=1}^{n} p_i = 1
\end{align*}

A maximization objective cannot serve as a loss function. Therefore, we convert the objective to the following minimization form to construct the unknown identity rejection loss function (denoted as $L_{uir}$):
\begin{align*}
L_{uir} = -\sum_{i=1}^{n} log(p_i)
\end{align*}

During the training process, the model is trained in a semi-supervised manner. We combine the fully-supervised loss function (denoted as $L_{sup}$) and our proposed semi-supervised loss function by weighted sum:
\begin{align*}
L_{overall} = L_{sup} + w \cdot L_{uir},
\end{align*}
where $w$ is a hyperparameter and we found $w=0.1$ is the best setup for our experiments through trial and error.

An intuitive perspective for observing the feature vectors and the weights in the fully-connected layer is to treat them as points distributed in the high dimensional space. It is assumed that features of the same class cluster around the corresponding center which is one of the row vectors in the fully-connected layer's weight matrix. Therefore, to enhance the discriminativeness of the model, centers for different identities must be distributed as sparsely as possible. When a model is well trained, for any of the samples, its feature vector is assumed to be able to approximate its center. Given a sample with an unknown identity, we denote its normalized feature vector as $f_u$, which should approximate the center of its corresponding identity. We denote the weight of the fully-connected layer as $W$ and the centers inside as $W_1, W_2, ..., W_n$. Then minimizing $f_u \cdot W_1, f_u \cdot W_2,..., f_u \cdot W_n$ can be considered as enlarging the distance between the center of $f_u$ and the centers of known identities. Hence it seems that the number of classes has been increased and extra space must be allocated to the new classes (represented by unlabeled samples), forcing the known classes to distribute more sparsely (see Fig. \ref{fig:figure1}). 

\subsection{Implementation Details}
\textbf{Numberical stability}: For a classification model that is capable of classifying a large amount of categories, since the activation values after softmax normalization sum up to 1, there can exist very small $p_i$'s. This can result in numerical unstability when calculating $log(p_i)$ for the UIR loss function. To overcome this problem, we applied one more softmax layer after these activation values. Since softmax will not change the relative magnitude relationships between activation values, this operation can avoid unstable computations while at the same time keep the classification result unchanged. We also found that this operation helps the model better converge to the optima.

\textbf{Identity overlap between labeled and unlabeled datasets}: Our assumption of the unlabeled dataset is that it contains no identities that appear in the labeled dataset. However, identity overlap can be inevitable, albeit it occurs infrequently. To remove the noisy data, we first use the face recognition model trained in a fully-supervised manner to do inference on these unlabeled data. We discard the images whose maximum activation value from the last fully connected layer is above a predefined threshold (0.9). We try to ensure that our assumption gets satisfied to the utmost extent. 

\textbf{Training strategy}: Since our goal of utilizing the unlabeled data is to further improve the performance of the trained model, there are two steps within our training process. Firstly we train the model in a fully-supervised manner until it converges. Then we deploy the UIR loss function along with the fully supervised loss function, training in a semi-supervised manner (See Fig. \ref{fig:figure2}). When we conduct the semi-supervised training, we keep the overall batch size unchanged, but decreasing the number of labeled images within a single batch to its 3/4 and supplement the rest 1/4 with unlabeled data. The two losses are calculated separately on these two portions of the batch data.

\begin{table*}[th!]
\centering
\caption{Performance (\%) of ResNet50 on the IJB-C benchmark.}
\label{tab:resnet-50-ijbc}
\begin{tabular}{cc|cccccc}
\hline
 &  & \multicolumn{6}{c}{TAR@FAR} \\ \hline
\multicolumn{1}{c|}{Methods} & model & $10^{-6}$ & $10^{-5}$ & $10^{-4}$ & $10^{-3}$ & $10^{-2}$ & $10^{-1}$ \\ \hline
\multicolumn{1}{c|}{\multirow{2}{*}{N0F0}} & baseline & 89.96 & 94.26 & 96.02 & \textbf{97.38} & \textbf{98.32} & 99.06 \\
\multicolumn{1}{c|}{} & ours & \textbf{90.41} & \textbf{94.33} & \textbf{96.03} & \textbf{97.38} & 98.30 & \textbf{99.08} \\ \hline
\multicolumn{1}{c|}{\multirow{2}{*}{N0F1}} & baseline & 90.53 & 94.42 & 96.03 & 97.38 & 98.30 & \textbf{99.12} \\
\multicolumn{1}{c|}{} & ours & \textbf{90.91} & \textbf{94.57} & \textbf{96.27} & \textbf{97.47} & \textbf{98.38} & \textbf{99.12} \\ \hline
\multicolumn{1}{c|}{\multirow{2}{*}{N1F0}} & baseline & 88.78 & 93.75 & \textbf{95.79} & \textbf{97.29} & \textbf{98.29} & 99.06 \\
\multicolumn{1}{c|}{} & ours & \textbf{88.81} & \textbf{93.86} & 95.78 & 97.27 & 98.27 & \textbf{99.07} \\ \hline
\multicolumn{1}{c|}{\multirow{2}{*}{N1F1}} & baseline & 89.13 & 93.92 & 95.99 & \textbf{97.39} & \textbf{98.40} & \textbf{99.11} \\
\multicolumn{1}{c|}{} & ours & \textbf{89.53} & \textbf{94.16} & \textbf{96.04} & 97.38 & 98.35 & 99.10 \\ \hline
\end{tabular}
\end{table*}

\begin{table*}[th!]
\centering
\caption{Performance (\%) of ResNet100 on the IJB-C benchmark.}
\label{tab:resnet-100-ijbc}
\begin{tabular}{cc|cccccc}
\hline
 &  & \multicolumn{6}{c}{TAR@FAR} \\ \hline
\multicolumn{1}{c|}{Methods} & model & $10^{-6}$ & $10^{-5}$ & $10^{-4}$ & $10^{-3}$ & $10^{-2}$ & $10^{-1}$ \\ \hline
\multicolumn{1}{c|}{\multirow{2}{*}{N0F0}} & baseline & 89.66 & 94.70 & 96.57 & 97.64 & \textbf{98.43} & 99.03 \\
\multicolumn{1}{c|}{} & ours & \textbf{91.46} & \textbf{94.96} & \textbf{96.70} & \textbf{97.68} & 98.42 & \textbf{99.06} \\ \hline
\multicolumn{1}{c|}{\multirow{2}{*}{N0F1}} & baseline & 89.07 & 94.80 & 96.77 & 97.78 & \textbf{98.48} & 99.11 \\
\multicolumn{1}{c|}{} & ours & \textbf{91.36} & \textbf{95.12} & \textbf{96.90} & \textbf{97.81} & 98.46 & \textbf{99.12} \\ \hline
\multicolumn{1}{c|}{\multirow{2}{*}{N1F0}} & baseline & 88.82 & 94.42 & 96.47 & 97.60 & \textbf{98.40} & 99.02 \\
\multicolumn{1}{c|}{} & ours & \textbf{90.35} & \textbf{94.60} & \textbf{96.55} & \textbf{97.65} & 98.38 & \textbf{99.04} \\ \hline
\multicolumn{1}{c|}{\multirow{2}{*}{N1F1}} & baseline & 88.42 & 94.45 & 96.66 & 97.71 & 98.43 & 99.07 \\
\multicolumn{1}{c|}{} & ours & \textbf{90.22} & \textbf{94.67} & \textbf{96.76} & \textbf{97.73} & \textbf{98.44} & \textbf{99.11} \\ \hline
\end{tabular}
\end{table*}

\begin{table*}[h!]
\centering
\caption{Performance (\%) of MobileNet-V1 on the IJB-C benchmark.}
\label{tab:MobileNet-V1-ijbc}
\begin{tabular}{cc|cccccc}
\hline
 &  & \multicolumn{6}{c}{TAR@FAR} \\ \hline
\multicolumn{1}{c|}{Methods} & model & $10^{-6}$ & $10^{-5}$ & $10^{-4}$ & $10^{-3}$ & $10^{-2}$ & $10^{-1}$ \\ \hline
\multicolumn{1}{c|}{\multirow{2}{*}{N0F0}} & baseline & 84.18 & 89.09 & 92.83 & 95.54 & \textbf{97.53} & \textbf{98.93} \\
\multicolumn{1}{c|}{} & ours & \textbf{84.69} & \textbf{89.50} & \textbf{92.95} & \textbf{95.55} & 97.49 & \textbf{98.93} \\ \hline
\multicolumn{1}{c|}{\multirow{2}{*}{N0F1}} & baseline & 84.61 & 89.44 & 93.16 & \textbf{95.87} & 97.69 & \textbf{99.03} \\
\multicolumn{1}{c|}{} & ours & \textbf{85.24} & \textbf{89.96} & \textbf{93.32} & 95.85 & \textbf{97.70} & 98.99 \\ \hline
\multicolumn{1}{c|}{\multirow{2}{*}{N1F0}} & baseline & 81.44 & 87.34 & 92.06 & 95.17 & \textbf{97.45} & 98.93 \\
\multicolumn{1}{c|}{} & ours & \textbf{82.03} & \textbf{88.07} & \textbf{92.26} & \textbf{95.20} & 97.44 & \textbf{98.94} \\ \hline
\multicolumn{1}{c|}{\multirow{2}{*}{N1F1}} & baseline & 80.02 & 87.63 & 92.27 & 95.53 & 97.59 & 99.04 \\
\multicolumn{1}{c|}{} & ours & \textbf{82.21} & \textbf{88.40} & \textbf{92.56} & \textbf{95.59} & \textbf{97.63} & \textbf{99.05} \\ \hline
\end{tabular}
\end{table*}

\begin{table*}[h!]
\centering
\caption{Performance (\%) of MobileNet-V2 on the IJB-C benchmark.}
\label{tab:MobileNet-V2-ijbc}
\begin{tabular}{cc|cccccc}
\hline
 &  & \multicolumn{6}{c}{TAR@FAR} \\ \hline
\multicolumn{1}{c|}{Methods} & model & $10^{-6}$ & $10^{-5}$ & $10^{-4}$ & $10^{-3}$ & $10^{-2}$ & $10^{-1}$ \\ \hline
\multicolumn{1}{c|}{\multirow{2}{*}{N0F0}} & baseline & 79.56 & 87.61 & 92.28 & 95.36 & 97.42 & \textbf{98.85} \\
\multicolumn{1}{c|}{} & ours & \textbf{79.77} & \textbf{87.96} & \textbf{92.36} & \textbf{95.44} & \textbf{97.44} & \textbf{98.85} \\ \hline
\multicolumn{1}{c|}{\multirow{2}{*}{N0F1}} & baseline & 79.85 & 88.03 & 92.68 & 95.63 & \textbf{97.53} & \textbf{98.91} \\
\multicolumn{1}{c|}{} & ours & \textbf{80.98} & \textbf{88.48} & \textbf{92.71} & \textbf{95.66} & \textbf{97.53} & 98.89 \\ \hline
\multicolumn{1}{c|}{\multirow{2}{*}{N1F0}} & baseline & 76.88 & 85.32 & 91.33 & 95.09 & 97.32 & \textbf{98.80} \\
\multicolumn{1}{c|}{} & ours & \textbf{76.98} & \textbf{86.09} & \textbf{91.56} & \textbf{95.17} & \textbf{97.37} & \textbf{98.80} \\ \hline
\multicolumn{1}{c|}{\multirow{2}{*}{N1F1}} & baseline & 75.66 & 85.91 & 91.70 & 95.35 & 97.45 & \textbf{98.90} \\
\multicolumn{1}{c|}{} & ours & \textbf{76.51} & \textbf{86.38} & \textbf{91.93} & \textbf{95.37} & \textbf{97.53} & 98.89 \\ \hline
\end{tabular}
\end{table*}

\section{Experiments}

\subsection{Datasets and Networks}
\textbf{Training datasets}: We use the currently largest public face recognition dataset MS-Celeb-1M \cite{msceleb1m} to serve as our labeled dataset. It contains about 100K identities and over 10M images. However, the original MS-Celeb-1M dataset suffers seriously from long-tailed data and noisy data. Therefore, we instead deploy MS1MV2 \cite{retinaface}, a refined version of MS-Celeb-1M. The refined dataset contains over 90K identities and 5 million images. To construct the unlabeled dataset, we crawled large amounts of face images from the Internet. After removing the face images with low quality or identity overlap with the labeled dataset, we obtained over 4.9 million images. All the face images are detected and preprocessed with RetinaFace \cite{retinaface} for the training process. 

\textbf{Testing Datasets}: We evaluate the effectiveness of our proposed approach on three large-scale benchmarks: IJB-C \cite{ijbc}, iQIYI-VID \cite{iqiyivid} and Trillion-Pairs \cite{trillionpairs}. For the IJB-C dataset, we validate the performance of our models on the 1:1 verification task, which contains 23,124 templates along with 19,557 genuine matches and 15,639,932 impostor matches. We report the True Accept Ratios (TAR) at different False Accept Ratios (FAR). For iQIYI-VID, we test the performance on the large-scale video test set, which includes 200K videos of 10K identities. We follow the protocol described in \cite{iccvchallenge} to assess the performance. The Trillion-Pairs dataset contains 1.58m images from Flickr as the gallery set and 274K images as the probe set. Again, we follow the protocol described in \cite{iccvchallenge} for evaluation. 

\textbf{Networks}: We deploy four popular networks to serve as the feature extractor, namely ResNet50, ResNet100 \cite{resnet}, MobileNet-V1 \cite{mobilenetv1} and MobileNet-V2 \cite{mobilenetv2}. The dimensionality of the feature is set to 512. Our implementation is based on the open source repository InsightFace \cite{insightfacecode}. For the supervised loss function, we use the state-of-the-art ArcFace loss function \cite{arcface}. We normalize the weigths in the last fully-connected layer and the features. Features are further rescaled to an amplitude of 64. These are the common practice in previous works \cite{arcface} \cite{sphereface}.

\subsection{Performance Evaluation}

We trained a total of 8 models for performance evaluation. For each of the four backbone networks, we train two models, namely the baseline model trained in a fully-supervised manner and our improved model further trained with unlabeled data by our proposed approach.

We firstly tested our models on the IJB-C benchmark. To construct the face representation, we explored two approaches for the postprocess of the extracted features, namely feature normalization and image flip, denoted as $N$ and $F$. Specifically for flip, we extract features from both the original and horizontally-flipped face images and aggregate the features by element-wise summation. Therefore, the combination of whether to apply feature normalization or image flip results in four methods for constructing the final representation (denoted as N0F0, N0F1, N1F0 and N1F1). 

From Tab. \ref{tab:resnet-50-ijbc} \ref{tab:resnet-100-ijbc} \ref{tab:MobileNet-V1-ijbc} \ref{tab:MobileNet-V2-ijbc} we can notice obvious improvement on the IJB-C benchmark under different backbone networks and different forms of face representation. Specifically for the ResNet100 network, we compare our performance with those reported in the work \cite{retinaface}, which deployed the same labeled dataset, supervised loss function and backbone network. At an FAR of $10^{-6}$, we achieved a TAR of 89.66\% with the vanilla features and a TAR of 91.36\% with flip, both outperforming the reported 88.73\% and 89.12\% in \cite{retinaface}.  

We also provide our experimental results on the TrillionPairs benchmark and the iQIYI-VID benckmark separately in Tab. \ref{tab:trip} and Tab. \ref{tab:iqiyi}. Here we construct the final representation with both flip and feature normalization. Compared to the baseline models, again we witness a considerable amount of performance gain. This indicates that features learned via our approach is indeed more discriminative than those from the baseline model. Our approach does enhanced the discriminativeness by utilizing the unlabeled data. 

\begin{table}[h!]
\centering
\caption{Performance Evaluation (\%) on the TrillionPairs benchmark.}
\label{tab:trip}
\begin{tabular}{cc|c}
\hline
 &  & TAP@FAR=$10^{-4}$ \\ \hline
\multicolumn{1}{c|}{\multirow{2}{*}{ResNet50}} & baseline & 90.041 \\
\multicolumn{1}{c|}{} & ours & \textbf{90.525} \\ \hline
\multicolumn{1}{c|}{\multirow{2}{*}{ResNet100}} & baseline & 92.955 \\
\multicolumn{1}{c|}{} & ours & \textbf{93.325} \\ \hline
\multicolumn{1}{c|}{\multirow{2}{*}{MobileNet-V1}} & baseline & 70.040 \\
\multicolumn{1}{c|}{} & ours & \textbf{71.176} \\ \hline
\multicolumn{1}{c|}{\multirow{2}{*}{MobileNet-V2}} & baseline & 68.699 \\
\multicolumn{1}{c|}{} & ours & \textbf{69.326} \\ \hline
\end{tabular}
\end{table}

\begin{table}[h!]
\centering
\caption{Performance Evaluation (\%) on the iQIYI-VID benchmark.}
\label{tab:iqiyi}
\begin{tabular}{cc|c}
\hline
 &  & TAP@FAR=$10^{-8}$ \\ \hline
\multicolumn{1}{c|}{\multirow{2}{*}{ResNet50}} & baseline & 60.228 \\
\multicolumn{1}{c|}{} & ours & \textbf{62.041} \\ \hline
\multicolumn{1}{c|}{\multirow{2}{*}{ResNet100}} & baseline & 65.139 \\
\multicolumn{1}{c|}{} & ours & \textbf{67.214} \\ \hline
\multicolumn{1}{c|}{\multirow{2}{*}{MobileNet-V1}} & baseline & 35.161 \\
\multicolumn{1}{c|}{} & ours & \textbf{37.656} \\ \hline
\multicolumn{1}{c|}{\multirow{2}{*}{MobileNet-V2}} & baseline & 28.780 \\
\multicolumn{1}{c|}{} & ours & \textbf{29.808} \\ \hline
\end{tabular}
\end{table}

\subsection{Further Analysis on Classification}

\textbf{Sparsity of the center distribution}: To improve the discriminativeness, we train the model to reject those faces with unknown identities. This process renders the distribution of the centers in the last fully-connected layer sparse (see Fig. \ref{fig:figure1}) so that samples with unknown idnetities will not be misclassified to any of the known identities. To illustrate this effect, we further probe into the sparsity of the distribution of the centers. We extract the row vectors (which represent centers of the corresponding identities) from the last fully-connected layer and calculate the pairwise distance between them. The distance metric we used here is the cosine distance:

\begin{equation*}
d_{cos}(\vec{u}, \vec{v}) = 1 - \frac{\vec{u} \cdot \vec{v}}{\|\vec{u}\| \cdot \|\vec{v}\|}
\end{equation*}

\begin{table*}[t]
\centering
\caption{Cumulative distribution of the activation values (softmax-normalized) from the `unknown' face images. }
\label{tab:c-dist}
\begin{tabular}{ll|ccccccccc}
\hline
 &  & \multicolumn{9}{c}{Cumulative distribution (\%) of the Activation Values} \\ \cline{3-11} 
 &  & \textless 0.1 & \textless 0.2 & \textless 0.3 & \textless 0.4 & \textless 0.5 & \textless 0.6 & \textless 0.7 & \textless 0.8 & \textless 0.9 \\ \hline
\multicolumn{1}{l|}{\multirow{2}{*}{ResNet50}} & baseline & 0.57 & 9.30 & 23.88 & 38.52 & 51.42 & 62.42 & 72.60 & 81.56 & 89.23 \\
\multicolumn{1}{l|}{} & ours & \textbf{0.63} & \textbf{10.06} & \textbf{25.74} & \textbf{40.50} & \textbf{53.77} & \textbf{64.46} & \textbf{74.43} & \textbf{82.32} & \textbf{89.90} \\ \hline
\multicolumn{1}{l|}{\multirow{2}{*}{ResNet100}} & baseline & 0.59 & 8.02 & 22.27 & 36.60 & 49.41 & 59.65 & 68.35 & 77.03 & 85.30 \\
\multicolumn{1}{l|}{} & ours & \textbf{1.28} & \textbf{11.42} & \textbf{26.55} & \textbf{41.54} & \textbf{53.64} & \textbf{63.56} & \textbf{72.23} & \textbf{79.91} & \textbf{87.51} \\ \hline
\multicolumn{1}{l|}{\multirow{2}{*}{MobileNet-V1}} & baseline & 0.52 & \textbf{9.74} & 25.80 & 41.33 & 54.52 & 65.69 & 74.87 & 83.15 & 90.32 \\
\multicolumn{1}{l|}{} & ours & \textbf{0.68} & 9.71 & \textbf{26.21} & \textbf{42.25} & \textbf{56.03} & \textbf{66.50} & \textbf{75.42} & \textbf{83.28} & \textbf{90.82} \\ \hline
\multicolumn{1}{l|}{\multirow{2}{*}{MobileNet-V2}} & baseline & 0.60 & 9.02 & 24.29 & 40.37 & 53.83 & 65.21 & 74.46 & 82.27 & 90.02 \\
\multicolumn{1}{l|}{} & ours & \textbf{0.75} & \textbf{9.61} & \textbf{25.18} & \textbf{41.00} & \textbf{54.88} & \textbf{65.90} & \textbf{74.63} & \textbf{82.71} & \textbf{90.38} \\ \hline
\end{tabular}
\end{table*}

\begin{table}[h!]
\centering
\caption{Average pairwise distances between centers (row vectors of the last fully connected layer). This distance indicates how sparse these centers are distributed. The more discriminative the representations are, the larger the average pairwise distance is.}
\label{tab:dist}
\begin{tabular}{l|l|l} 
\hline
             & \multicolumn{2}{l}{Average Pairwise Distance }  \\ 
\hline
Networks     & baseline & ours                             \\ 
\hline
MobileNet-V2 & 0.998089 & \textbf{0.998109}                             \\ 
\hline
MobileNet-V1 & 0.998765 & \textbf{0.998784}                              \\
\hline
ResNet50     & 1.000923 & \textbf{1.000937}                             \\ 
\hline
ResNet100    & 1.001652 & \textbf{1.001657}                             \\ 
\hline

\end{tabular}
\end{table}

Tab. \ref{tab:dist} lists the average pairwise distances between centers. Note that the more discriminative the representation is, the larger the pairwise distances are (see the baseline column of Tab. \ref{tab:dist}). We can notice that by further training the model with unlabeled dataset, the ours model has a larger pairwise distance between centers, indicating that the distribution of the samples are distributed more sparsely. 

\textbf{UIR loss can improve the open-set recognition performance}: Open set recognition categorize objects into two sets: the \textit{known} set and the \textit{unknown} set. It aims at accurately recognizing objects that belongs to the known set and rejecting those belonging to the unknown set. Specifically for face recognition, it is also an open-set problem where `unknown' identities must be rejected. A common practice for this problem is to threshold the maximal activation value which serves as a confidence score with a predefined value. However sometimes samples with unknown identities can also produce large activation values. Here we show that our UIR loss function can improve the open-set recognition performance by suppressing the activation values to some extent.

To illustrate this, we collected extra 10K face images whose identities are not included in our training set and preprocessed them in the same way as we processed our unlabeled training data. We fed them into the networks and collect statistics from their final activation values for classification. Tab. \ref{tab:c-dist} shows the cumulative distribution of the maximal activation values over the 10K images. It lists the amount of activations of the 10K face images with unknown identities under various thresholds. We can notice that under a certain threshold, the percentage of the activation values for our improved model is always higher than that of the baseline model. This shows the improvement on the ability of rejecting unknown samples brought by our UIR loss.

\begin{table}[h!]
\centering
\caption{Average activation values (softmax-normalized) of the 10K face images with unknown identities. The lower the value is, the better. }
\label{tab:act-rej}
\begin{tabular}{l|l|l} 
\hline
             & \multicolumn{2}{l}{Average Activation Value }  \\ 
\hline
Networks     & baseline & ours                     \\ 
\hline
ResNet50 & 0.5226 & \textbf{0.5098}                \\ 
\hline
ResNet100 & 0.5454 & \textbf{0.5141}               \\
\hline
MobileNet-V1    & 0.5054 & \textbf{0.5002}         \\ 
\hline
MobileNet-V2    & 0.5115 & \textbf{0.5064}         \\ 
\hline

\end{tabular}
\end{table}

We also list the average values for these activations in Tab. \ref{tab:act-rej}. We can notice an obvious drop of the average activation value in our improved model compared with the baseline. This also indicates that our UIR loss function can improve the open-set recognition performance to some extent.

\section{Conclusion}

In this paper, we proposed a novel Unknown Identity Rejection (UIR) loss that utilizes unlabeled data to further enhance the discriminativeness of the learned feature representation. We provide both mathematical explanation and geometrical interpretation of our proposed UIR loss. Moreover, extensive experiments have illustrated the effectiveness of out approach. UIR loss is simple and easy to be deployed to many other frameworks. Since most of the face representation learning tasks are based on the classification of different identities, we believe that our approach is also possible to be applied to other classification-based tasks. For future work, we will conduct experiments on other classification-based deep learning tasks, such as person re-identification, fine-grained image recognition, etc. Moreover, based on the improvement on open-set face recognition, we will also probe into possible means to further suppress the activation values of the images with unknown categories. We hope that more potentials can be further exploited from our proposed approach.

\newpage

{\small
\bibliographystyle{ieee_fullname}
\bibliography{egbib}
}

\end{document}